\documentclass[runningheads]{llncs}

\usepackage{eccv}

\usepackage{eccvabbrv}
\usepackage{overpic}
\usepackage{graphicx}
\usepackage{booktabs}
\usepackage{enumitem}
\usepackage{threeparttable}
\usepackage{tabularx}
\usepackage{array}
\usepackage{float}
\usepackage[accsupp]{axessibility}  
\usepackage{bibunits}
\defaultbibliographystyle{splncs04}

\usepackage{hyperref}

\usepackage{orcidlink}

\makeatletter
\renewcommand\section{\@startsection{section}{1}{\z@}%
  {-1.6ex \@plus -0.4ex \@minus -0.2ex}%
  {0.8ex \@plus 0.2ex}%
  {\normalfont\large\bfseries}}

\renewcommand\subsection{\@startsection{subsection}{2}{\z@}%
  {-1.2ex \@plus -0.3ex \@minus -0.2ex}%
  {0.6ex \@plus 0.2ex}%
  {\normalfont\normalsize\bfseries}}

\renewcommand\subsubsection{\@startsection{subsubsection}{3}{\z@}%
  {-1.0ex \@plus -0.3ex \@minus -0.2ex}%
  {0.5ex \@plus 0.2ex}%
  {\normalfont\normalsize\bfseries}}
\renewcommand\paragraph{\@startsection{paragraph}{4}{\z@}%
  {-0.8ex \@plus -0.2ex \@minus -0.1ex}%
  {0.4em}%
  {\normalfont\normalsize\bfseries}}

\renewcommand\subsubsection{\@startsection{subsubsection}{3}{\z@}%
  {-1.0ex \@plus -0.3ex \@minus -0.2ex}%
  {-0.5em}%
  {\normalfont\normalsize\bfseries}}

\newcommand{\samethanks}{\@thanks}

\makeatother

\begin{document}

\title{Giving Faces Their Feelings Back}
\subtitle{\small{Explicit Emotion Control for Feedforward Single-Image 3D Head Avatars}}

\titlerunning{Giving Faces Their Feelings Back}

\author{Yicheng Gong\inst{1,2} \and
Jiawei Zhang\inst{1,2} \and
Liqiang Liu\inst{1,3} \and
Yanwen Wang\inst{1,2} \and
Lei Chu\inst{2} \and
Jiahao Li\inst{2} \and
Hao Pan\inst{4} \and
Hao Zhu\inst{1} \and
Yan Lu\inst{2}
} 
\authorrunning{Y. Gong et al.}

\institute{Nanjing University, Nanjing, China\\ 
\and
Microsoft Research Asia, Beijing, China\\
\and
University of Science and Technology of China, Hefei, China\\
\and
Tsinghua University, Beijing, China\\
}

\maketitle
\begin{bibunit}
\begin{figure*}[htbp]
\vspace{-3mm}
  \centering
  \includegraphics[width=1.0\textwidth]{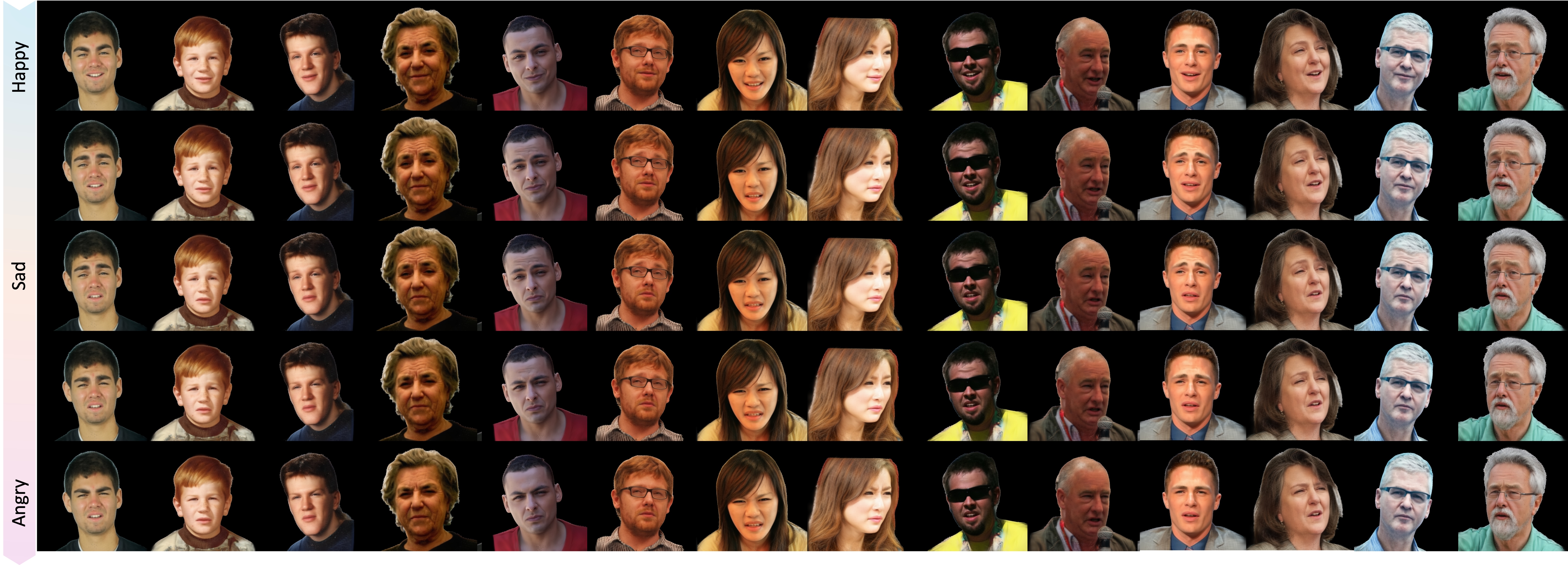}
  \vspace{-4mm}
  \caption{Emotion-Interpolated 3D Head Avatars. 
Given a single image, our feed-forward framework reconstructs expressive 3D avatars with explicit and interpolatable emotion control. 
The same blended emotion exhibits identity-aware variations, and all results are produced in a single forward pass without per-identity optimization.}
  \vspace{-12.5mm}
  \label{fig:teaser}
\end{figure*}

\begin{abstract}
We present a framework for explicit emotion control in feed-forward, single-image 3D head avatar reconstruction. Unlike existing pipelines where emotion is implicitly entangled with geometry or appearance, we treat emotion as a first-class control signal that can be manipulated independently and consistently across identities. Our method injects emotion into existing feed-forward architectures via a dual-path modulation mechanism without modifying their core design. Geometry modulation performs emotion-conditioned normalization in the original parametric space, disentangling emotional state from speech-driven articulation, while appearance modulation captures identity-aware, emotion-dependent visual cues beyond geometry. To enable learning under this setting, we construct a time-synchronized, emotion-consistent multi-identity dataset by transferring aligned emotional dynamics across identities. Integrated into multiple state-of-the-art backbones, our framework preserves reconstruction and reenactment fidelity while enabling controllable emotion transfer, disentangled manipulation, and smooth emotion interpolation, advancing expressive and scalable 3D head avatars.
\end{abstract}

\keywords{Single-image 3D head avatars; Emotion-controllable facial animation; Feed-forward reconstruction; FLAME; 3D Gaussian Splatting}

\section{Introduction}
High-quality 3D head avatars reconstructed from a single image have become a key component of virtual humans, telepresence, and embodied agents. Recent feed-forward pipelines recover identity-consistent geometry and appearance from a single RGB image, enabling scalable avatar creation without subject-specific optimization. Despite this progress, emotional expressiveness remains limited. In most existing systems, emotion is not modeled as an explicit control variable but is implicitly entangled within facial expression parameters or driving signals, making it difficult to manipulate emotion in a controllable and identity-consistent manner.

A core challenge lies in the relationship between emotion and facial articulation. Parametric models such as FLAME~\cite{flame} encode speech-driven motion, co-articulation, and residual emotional influence within a shared deformation space. While expressive, this representation conflates multiple factors and therefore lacks a clean control axis for emotion, limiting stable and interpretable manipulation. Moreover, emotion is not conveyed through geometry alone: subtle appearance cues, such as wrinkles, shading, and skin tension, contribute significantly to perceived affect and interact strongly with identity. The same emotional state may manifest differently across individuals due to identity-specific facial characteristics. These observations indicate that emotion should be modeled as an identity-aware control signal that jointly modulates geometry and appearance.

In this work, we introduce a framework for explicit emotion control in single-image 3D head avatar reconstruction. Our key insight is to elevate emotion to a first-class control variable that can be integrated into existing feed-forward pipelines without altering their core structure. We propose a dual-path emotion-aware modulation mechanism operating within the original parameterization. Emotional variation is decomposed into complementary geometry and appearance pathways: geometry modulation reconditions expression and jaw parameters based on emotion, while appearance modulation captures identity-dependent visual changes beyond geometry. Importantly, geometry modulation does not introduce new deformation bases or increase the degrees of freedom. Instead, it acts as an emotion normalization mechanism in parameter space. By conditioning expression parameters on an explicit emotion signal, the modulation removes residual emotional bias embedded in motion sequences, often coupled with speech articulation, and aligns geometry with a target emotional state. This normalization enables emotion to be controlled independently of speech dynamics while preserving identity and articulation fidelity.

Our design is plug-and-play at the architectural level, operating directly on the interaction between geometry queries and appearance representations. It is therefore compatible with a wide range of feed-forward reconstruction pipelines, enabling explicit emotion control without degrading reconstruction quality or increasing architectural complexity. This setting differs fundamentally from prior emotion-aware avatar synthesis approaches that rely on identity-specific optimization or calibrated multi-view capture. For example, methods such as EmoTalk3D~\cite{emotalk3d} achieve high-fidelity emotional animation through per-identity training but do not generalize emotion control to unseen identities from a single image in a purely feed-forward manner. In contrast, we target identity-agnostic emotion modulation that generalizes across subjects without subject-specific optimization.

Learning such controllable avatars requires appropriate supervision. Existing datasets either provide identity-specific emotion annotations or entangle emotional variation with identity and motion, hindering factor isolation. To address this, we construct a time-synchronized, multi-view facial dataset with explicit emotion annotations. By transferring synchronized emotional dynamics from a small set of anchor subjects to a large pool of identity images, the dataset preserves temporal emotion structure while decoupling it from identity. This design enables systematic analysis of emotion-induced geometry and appearance variation and provides the supervision necessary for identity-agnostic emotion control.

We integrate our framework into multiple state-of-the-art single-image avatar reconstruction pipelines. Experiments demonstrate improved emotional expressiveness, identity consistency, and controllability without compromising reconstruction fidelity. Ablation and analysis show that both geometry and appearance modulation are necessary for convincing emotion control, and that geometry normalization plays a critical role in decoupling emotional state from speech-driven motion.

In summary, this work makes the following contributions:
\begin{itemize}[leftmargin=*, itemsep=0pt, topsep=2pt]
    \item We introduce \emph{emotion as an explicit, first-class control dimension} in feed-forward 3D head avatar reconstruction from a single image, establishing the first framework that enables identity-consistent and controllable emotional variation \emph{without subject-specific optimization or iterative fitting}.
    \item We propose a dual-path emotion-aware modulation mechanism that factorizes emotional influence into complementary geometry and appearance pathways: geometry modulation performs emotion normalization within the original parametric space, while appearance modulation models identity-dependent, emotion-specific visual responses beyond geometry.
    \item We construct a time-synchronized, emotion-consistent multi-identity facial dataset that isolates emotional variation from speech-driven articulation and identity, enabling systematic learning and analysis of emotion-controllable avatar reconstruction in a feed-forward setting.
\end{itemize}

We believe this work represents a principled step toward emotionally expressive, controllable, and scalable 3D avatars suitable for next-generation virtual humans and embodied agents. The code will be released upon acceptance.
\begin{figure*}[t]
\vspace{-3mm}
  \centering
    \includegraphics[width=\linewidth]{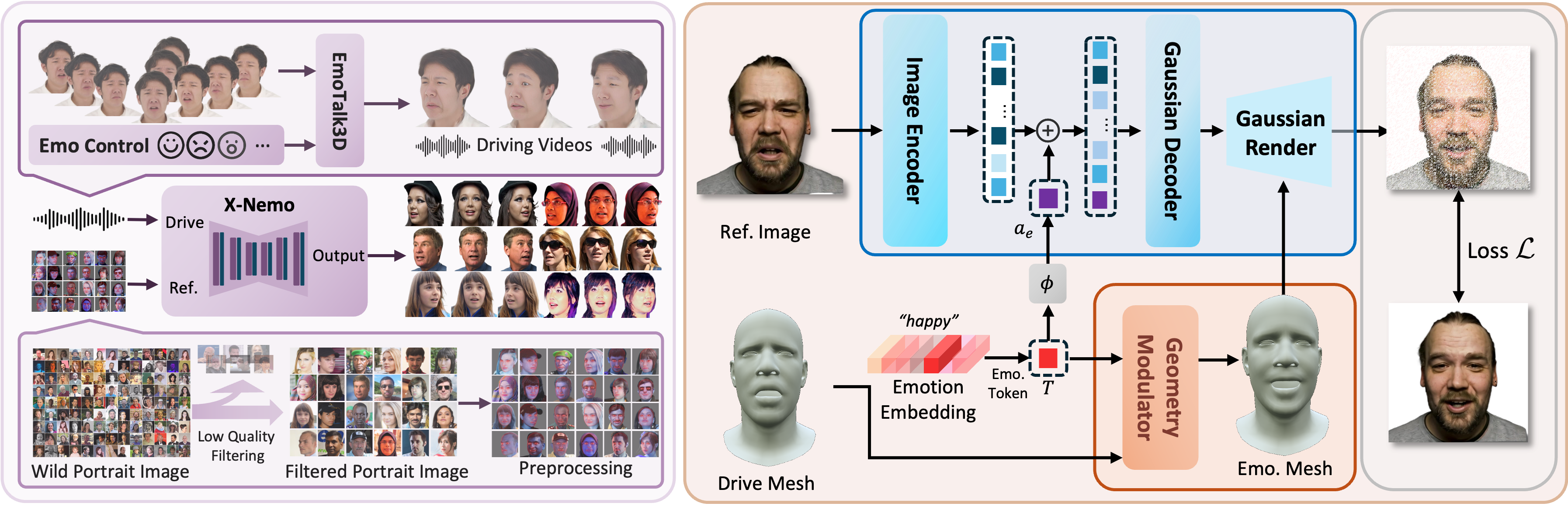}
    \vspace{-5mm}
\caption{Overview of the proposed framework. 
The system consists of (left) emotion-consistent data curation and (right) dual-path emotion-aware modulation. 
We build a time-synchronized multi-identity dataset by transferring frame-aligned emotional dynamics from anchor subjects, yielding explicit emotion supervision disentangled from speech and identity. 
The reconstruction network modulates geometry and appearance conditioned on a single reference image, driving FLAME parameters, and an emotion label, enabling controllable and identity-consistent emotional synthesis.}
\vspace{-8mm}
  \label{fig:pipeline}
\end{figure*}

\section{Related Work}
\subsection{3D Head Modeling}


Single-image 3D head modeling methods fall into person-specific optimization and generalized feed-forward reconstruction.
Person-specific pipelines recover animatable avatars from limited observations using 3DMM~\cite{3dmm} / FLAME~\cite{flame}, often with implicit neural representations~\cite{nerf, headnerf, insta, bhattarai2024triplanenet}.
Recently, 3D Gaussian Splatting~\cite{3dgs} enables real-time, high-fidelity dynamic head avatars via compact Gaussian fields and deformation models~\cite{flashavatar, gaussianavatars, monogaussianavatar, gsheadavatar, streamme, hravatar, rgbavatar, wang2024gaussianhead, gsheadshoulder, zhang2025fate, dhamo2024headgas}.


Generalized approaches aim to directly regress animatable avatars from minimal input without per-subject optimization~\cite{Gao2022nerfblendshape, buehler2021varitex, kirschstein2023nersemble, zhuang2022mofanerf, MoRF, AVA, zhang2025bringingportrait3dpresence, liu2024headartist, liu2025avatarartist, li2023one, li2024generalizable}.
Some leverage multi-view datasets~\cite{kirschstein2023nersemble, ava256, FaceWarehouse, yang2020facescape, wang2022faceverse, li2024topo4d}in a data-driven manner, while others learn from in-the-wild images via generative model~\cite{Chan2022EG3D, piGAN2021, sun2023next3d, xie2022hfgi3d, 3davatargan, buhler2023preface, giebenhain2024npga, Gao2024PortraitGen, Gao2025Learn2Control, li2026condition, li2025hyplanehead, gu2025diffportrait360, he2025head}.
Recent feed-forward methods predict canonical Gaussian or mesh-aligned representations for immediate reenactment~\cite{gagavatar, lam, wu2026uika, ji2026fastgha}, while stronger priors and supervision further improve robustness across identities~\cite{portrait4d, portrait4dv2, GSM, kirschstein2024gghead, li2024spherehead, flexavatar, gpavatar, zhao2024invertavatar, avat3r, mvp4d, cap4d, li2025textalker}.
Despite these advances, emotion is typically treated implicitly, limiting explicit and controllable emotional modeling.

\subsection{Emotional 3D Head Modeling}

Emotion-controllable 3D head modeling is constrained by the limited availability of identity-diverse, temporally aligned emotional 3D data.
Datasets such as MEAD~\cite{mead} and EmoTalk3D~\cite{emotalk3d} provide valuable emotion supervision, but large-scale data suitable for feed-forward, identity-generalized emotion control remain scarce.

Most emotion-aware face generation methods operate in 2D~\cite{evp, ned, eamm, tan2025edtalk, EmoTalker, emmn, hong2022depth, ma2023otavatar, mir2023dithead, prajwal2020wav2lip, stypulkowski2024diffused, wang2021one, wang2023progressive, wei2024aniportrait, hallo, hallo2, hallo3, hallo4, xu2024vasa, ye2024real3d, yin2022styleheat, yu2023thpad, zhang2021flow, zhang2023metaportrait, zhou2020makeittalk, li2026personalive}, conditioning on emotion labels or inferring affect implicitly from audio--visual cues~\cite{facexhubert, xing2023codetalker, faceformer2022, sun2024diffposetalk, richard2021meshtalk, zhang2023sadtalker}.
In contrast, emotion modeling in 3D has been explored mainly in identity-specific or audio-driven settings~\cite{emotalk3d, emodifftalk, Ma2026ESGaussianFace, vasa3d, aneja2025gaussianspeech}, where recent results suggest that convincing emotion depiction requires modeling both geometric deformation and appearance variation.
However, existing approaches do not support explicit, feed-forward emotion control that generalizes across identities from a single image, leaving this setting insufficiently addressed.

\section{Method}
\subsection{Problem Setup and Preliminaries}

We study emotion-controllable 3D head avatar reconstruction in a feed-forward setting. 
Given a reference facial image, a sequence of parametric facial motion coefficients, and an explicit emotion signal, the goal is to reconstruct a renderable 3D head avatar whose geometry and appearance reflect the specified emotion while preserving identity and articulation. Formally, we define
\begin{equation}
   \mathcal{A} = \mathcal{F}(I, \mathbf{p}_{1:T}, e),
\end{equation}
where $I \in \mathcal{I}$ is a reference RGB image encoding identity-specific appearance, 
$\mathbf{p}_{1:T} = \{\mathbf{p}_t\}_{t=1}^{T}$ is a time-varying sequence of parametric head geometry coefficients serving as the driving signal, and 
$e \in \mathcal{E}$ is a discrete emotion label independent of the driving motion. 
The output $\mathcal{A}$ denotes a renderable avatar representation that can be animated by $\mathbf{p}_{1:T}$ while exhibiting emotion-consistent geometric and appearance variations.

\subsubsection{Baseline Feed-Forward Reconstruction.}
Modern single-image 3D head reconstruction pipelines can be abstracted as a feed-forward mapping
\begin{equation}
\mathcal{F}_0 : (I, \mathbf{p}_{1:T}) \mapsto \mathcal{A},
\end{equation}
where reconstruction arises from learned interactions between 
(i) \emph{geometry queries} derived from the driving parametric geometry $\mathbf{p}_{1:T}$ and 
(ii) \emph{appearance representations} extracted from the reference image $I$, 
typically implemented via cross-attention or feature modulation.

In such pipelines, emotion is not explicitly modeled. Any affective variation emerges implicitly from correlations within the geometry parameters or appearance features. Since $\mathbf{p}_{1:T}$ encodes speech-driven articulation together with residual affective bias, and appearance features capture identity without explicit emotion conditioning, emotion becomes entangled with motion and identity, hindering stable and interpretable emotional control.

\subsection{Emotion-Aware Modulation of Geometry and Appearance}

The overall pipeline of our method is illustrated in Fig. \ref{fig:pipeline}. To enable explicit and controllable emotional variation, we introduce a \emph{dual-path emotion-aware modulation} framework that injects emotion as an explicit conditioning signal into the reconstruction process. Emotional influence is decomposed into two complementary pathways: geometry modulation and appearance modulation. This design reflects the observation that emotion manifests through both geometric deformation and identity-dependent appearance variation.

\subsubsection{Emotion Representation.}

Let $e \in \mathbb{R}^{1 \times N}$ be a one-hot emotion label over $N$ categories. We define a learnable embedding table
\begin{equation}
E \in \mathbb{R}^{D \times N},
\end{equation}
and construct the masked emotion token
\begin{equation}
T = E \cdot \mathrm{diag}(e).
\end{equation}
The token $T \in \mathbb{R}^{D \times N}$ serves as the sole explicit emotion input for both geometry and appearance modulation. It is designed to be \emph{identity-independent}: it does not encode subject-specific appearance or geometry, and is shared across all identities.

\subsubsection{Emotion-Aware Geometry Modulation.}
The geometry pathway conditions the driving FLAME parameters on the emotion token. Given expression and jaw parameters $\mathbf{p}_t = (\mathbf{p}_t^{\mathrm{exp}}, \mathbf{p}_t^{\mathrm{jaw}})$, we apply an emotion-conditioned transformation
\begin{equation}
\tilde{\mathbf{p}}_t = g(\mathbf{p}_t, T),
\end{equation}
where $g(\cdot)$ denotes a learnable modulation function which we implemented via cross-attention.

This transformation preserves the original FLAME parameter space and does not introduce additional deformation bases or degrees of freedom. Instead, it acts as an \emph{emotion-conditioned normalization} in parameter space. Since expression parameters often contain residual affective bias coupled with speech-driven articulation, conditioning on $T$ removes such bias and aligns the geometry with the target emotion while retaining articulation fidelity. This enables stable emotion control independent of speech dynamics.

\subsubsection{Emotion-Aware Appearance Modulation.}

While geometry modulation accounts for deformation, many emotional cues arise from identity-specific appearance variations such as wrinkles,  shading and variations in skin tension. To capture these effects, we introduce an appearance modulation pathway.

Let $\mathbf{a}$ denote appearance features extracted from the reference image $I$. We generate emotion-aware appearance tokens $\mathbf{a}_e = \phi(T)$ and form
\begin{equation}
\tilde{\mathbf{a}} = [\mathbf{a} \;\| \; \mathbf{a}_e].
\end{equation}

Unlike geometry modulation, appearance modulation is inherently identity-aware: the emotion tokens modulate the identity-specific representation rather than prescribing fixed visual templates. It produces emotion-dependent visual variations consistent with the subject’s facial characteristics. This allows the same emotional state to manifest differently across identities, reflecting natural inter-subject variation. The resulting $\tilde{\mathbf{a}}$ is used in the geometry–appearance interaction module to produce the final avatar representation.

Together, the two pathways enable explicit, flexible, and identity-consistent emotion control while remaining fully compatible with existing feed-forward reconstruction architectures.

\subsection{Emotion-Consistent Multi-Identity Data Curation}
\label{sec:data_curation}

Learning explicit emotion control requires supervision that disentangles \emph{emotion}, \emph{speech-driven articulation}, and \emph{identity}. Existing datasets often entangle these factors: emotion annotations are identity-specific, temporally misaligned across takes, or coupled with motion and appearance variations. To address this limitation, we construct an \emph{emotion-consistent, time-synchronized, multi-identity} dataset tailored for feed-forward emotion control.

\subsubsection{Emotion-synchronized anchor subjects.}
We first build a small set of anchor sequences with temporally aligned emotional dynamics. Using EmoTalk3D~\cite{emotalk3d}, we generate multi-view sequences for an anchor subject under multiple discrete emotions.
This anchor subject satisfies a strict property:
\emph{the same person utters the same sentence with identical timing across different emotions}. Consequently, all emotion sequences for an anchor subject are frame-synchronized, share identical phoneme timing and articulation, and differ primarily in emotion-induced facial behavior.
This design yields frame-wise correspondence across emotions, allowing geometry and appearance differences to be attributed predominantly to emotional variation rather than speech or timing drift.
These anchor sequences provide the core supervision signal that disentangles emotion from articulation and serves as a canonical reference for emotion dynamics.

\subsubsection{Identity source images.}
To scale supervision across identities, we curate a large pool of high-quality identity images (one per identity), filtering out extreme poses and occlusions to ensure stable reconstruction. These images serve purely as \emph{identity and appearance sources} and are not required to contain any explicit emotional content.
This separation ensures that identity-specific appearance characteristics are preserved independently of the emotional motion source.

\subsubsection{Emotion transfer across identities.}
We transfer synchronized emotional dynamics from anchor sequences to diverse identities using a 2D reenactment model (X-NeMo~\cite{zhao2025xnemo}). For each anchor sequence, we animate each identity image while preserving articulation timing and emotion labels. This process generates facial video sequences where:
(i) articulation and temporal structure are inherited from the anchor,
(ii) emotion labels remain unchanged, and
(iii) identity-specific geometry and appearance are preserved from the target image.
Importantly, while the emotional state is shared, the resulting appearance variations are allowed, and expected, to differ across identities, capturing natural inter-subject variability such as different wrinkle patterns or skin responses under the same emotion. 

\subsubsection{Resulting supervision and necessity.}
The final dataset provides multi-view, emotion-labeled and time-synchronized sequences across massive identities, capturing two complementary variations:
(1) emotion-induced geometry differences under identical articulation, enabled by anchor synchronization; and 
(2) identity-aware appearance responses to emotions, enabled by large-scale identity transfer.
This supervision enables learning emotion normalization in geometry space without conflating emotion with speech motion, and learning appearance modulation where the same emotion manifests differently across identities. Without synchronized and identity-diverse supervision, emotion control would remain entangled with articulation or collapse into identity-specific patterns, hindering reliable feed-forward generalization.

\section{Experiments}
\subsection{Experimental Setup}
\subsubsection{Backbones.}
We evaluate our approach on two representative feed-forward 3D portrait/head reconstruction frameworks: \textit{LAM}~\cite{lam} and \textit{Zhang et al.}~\cite{zhang2025bringingportrait3dpresence}. Our emotion-aware training and data pipeline is applied on top of their public implementations or equivalent setups, demonstrating plug-and-play compatibility across different backbones.

\textit{LAM}~\cite{lam} predicts renderable Gaussian attributes by conditioning canonical head representations (e.g., FLAME~\cite{flame}) on a reference image and reenacts motion via driving sequences. We fine-tune from the released checkpoint.   \textit{Zhang et al.}~\cite{zhang2025bringingportrait3dpresence} project image features into a canonical UV representation and predict Gaussian attributes driven by a target mesh; we follow their training protocol and replace the template mesh with FLAME for compatibility. We retrained the model of Zhang et al. on our dataset, as their original checkpoints were not publicly released.

\subsubsection{Implementation.}
For fair comparison, we retain each backbone’s original training settings, including preprocessing, rendering resolution, optimizers, and learning-rate schedules.

Emotion labels are encoded as \textit{continous} embeddings and concatenated with image tokens to condition the Transformer. Geometry modulation is implemented using a lightweight Transformer operating on FLAME expression and jaw parameters.

All evaluation identities and sequences are strictly held out from training, and anchor synchronization sequences are excluded from evaluation. Detailed dataset statistics, hyper-parameters, objectives and more results are provided in the supplementary material.

\subsection{Evaluation Protocols}

We evaluate the proposed framework from three aspects: 
\emph{(i) reconstruction and reenactment fidelity}, 
\emph{(ii) emotion transfer and disentanglement}, and 
\emph{(iii) perceptual emotion fidelity}. 
Quantitative evaluation is conducted on paired synthetic data where ground truth is available, while real-image analysis and user studies are used for qualitative assessment. 
All evaluation identities and sequences are strictly held out from training, and anchor synchronization sequences are excluded.

\subsubsection{Reconstruction and Reenactment Fidelity.}
To verify that emotion-aware modulation does not degrade backbone performance, we evaluate on a held-out synthetic dataset with pixel-aligned ground truth. 
We consider both self- and cross-reenactment settings and report PSNR, SSIM, and LPIPS for image fidelity, CSIM (ArcFace cosine similarity) for identity consistency, and AED and APD for geometric driving accuracy. 
These metrics assess reconstruction fidelity and motion preservation. Additional real-image qualititive results are provided in the supplemental material.

\begin{figure}[!t]
  \centering
    \begin{overpic}[width=0.8\linewidth]{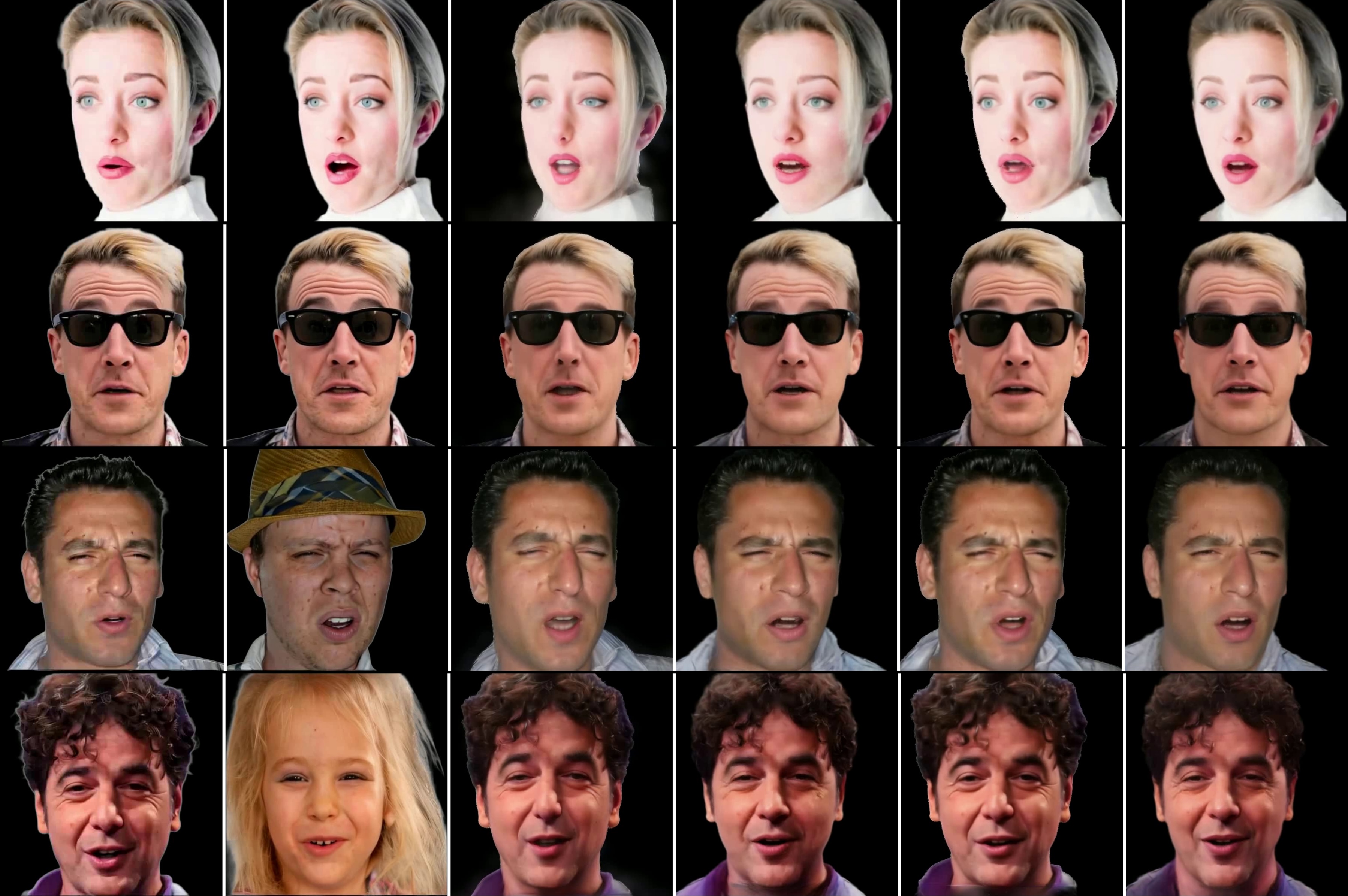}
    \put(5.5,-3){\footnotesize Ref.}
    \put(21.5,-3){\footnotesize Drv.}
    \put(37.8,-3){\footnotesize LAM}
    \put(53.3,-3){\footnotesize Zhang}
    \put(70.9,-3){\footnotesize LAM$^\dagger$}
    \put(86.5,-3){\footnotesize Zhang$^\dagger$}

    \put(-3.5,37){\rotatebox{90}{\footnotesize Self-reenactment}}
    \put(-3.5,3){\rotatebox{90}{\footnotesize Cross-reenactment}}
    \end{overpic}
\caption{Reconstruction and reenactment on a held-out synthetic dataset. 
Top: self-identity reenactment using the subject's own motion. 
Bottom: cross-identity reenactment driven by another subject.
Compared to the original baseline, emotion-aware modulation introduces no degradation in generation quality, consistently maintaining the same level of reconstruction fidelity and driving accuracy across settings and backbones.}
  \label{fig:recon_fidelity}
  \vspace{-6mm}
\end{figure}
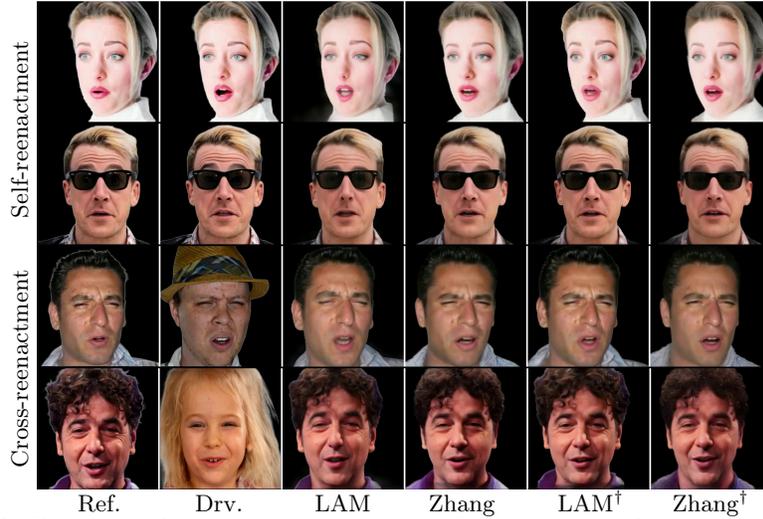

\subsubsection{Emotion Transfer and Disentanglement.}
Emotion transfer is evaluated on the same synthetic test set, where ground truth under different target emotions enables pixel-level comparison. 
We additionally provide qualitative results on both synthetic and real images (all unseen during training), including rendered avatars and corresponding FLAME geometry to visualize emotion-induced changes in deformation and appearance.

\subsubsection{User Study.}
We conduct a user study on real images to evaluate emotion recognizability, naturalness, and rendering quality, complementing the quantitative results.

\begin{figure}[t]
  \centering
    \begin{overpic}[width=0.9\linewidth]{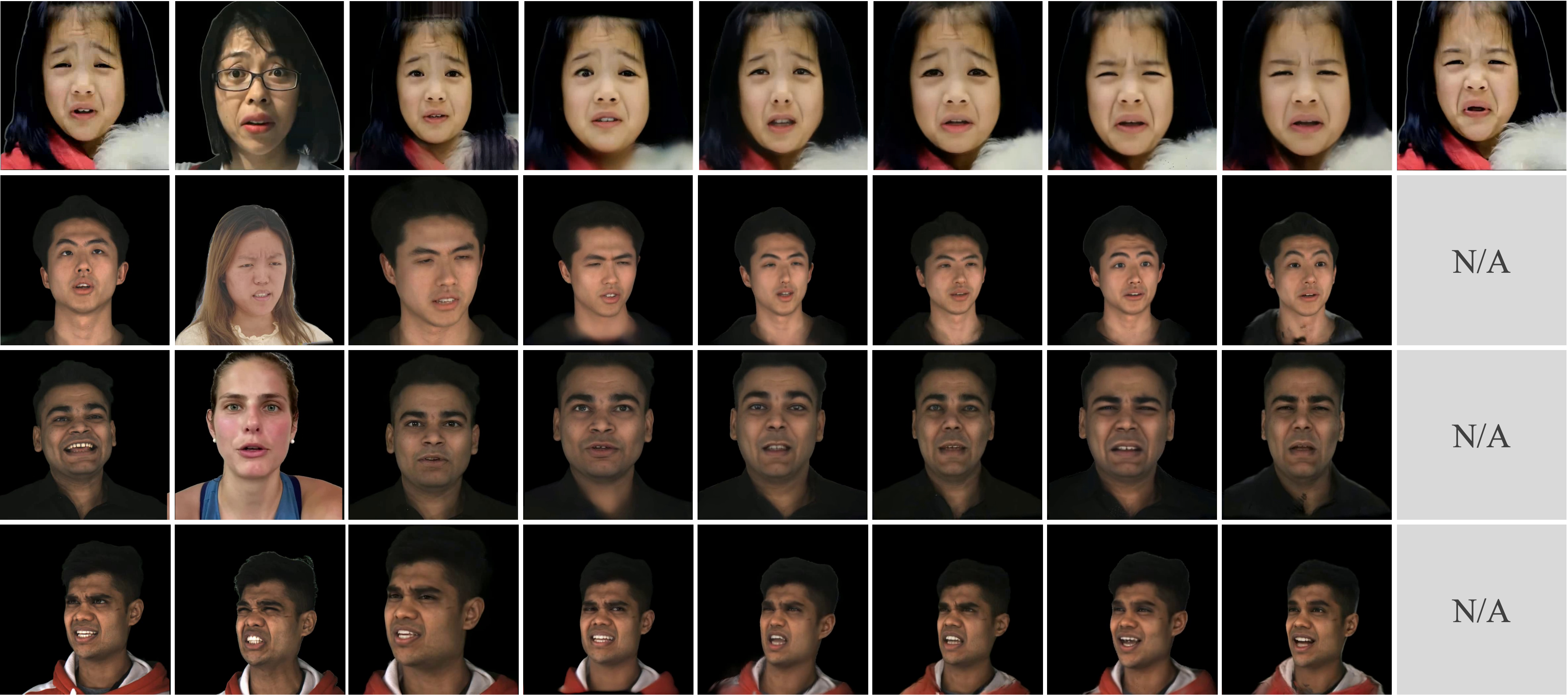}
    \put(3,-2.7){\footnotesize Ref.}
    \put(14,-2.7){\footnotesize Drv.}
    \put(23,-2.7){\footnotesize P4D-v2}
    \put(35.5,-2.7){\footnotesize GAG}
    \put(46.5,-2.7){\footnotesize LAM}
    \put(57.2,-2.7){\footnotesize Zhang}
    \put(68.3,-2.7){\footnotesize LAM$^\dagger$}
    \put(78.8,-2.7){\footnotesize Zhang$^\dagger$}
    \put(92,-2.7){\footnotesize GT}

    \put(-3.5,1){\rotatebox{90}{\footnotesize Happy}}
    \put(-3.5,11.5){\rotatebox{90}{\footnotesize Disgust}}
    \put(-3.5,24.5){\rotatebox{90}{\footnotesize Fear}}
    \put(-3.5,34){\rotatebox{90}{\footnotesize Disgust}}
    \end{overpic}
  \caption{Emotion transfer with explicit emotion control. 
Rows show four identities (top: synthetic with GT; others: real), and columns compare methods. 
The target emotion is indicated on the left. 
Our approach enforces the specified emotion independent of source appearance or motion, yielding identity-consistent results.}
  \label{fig:emotion_transfer}
   \vspace{-3mm}
\end{figure}

\begin{table}[!h]
\vspace{-6mm}
\centering
\scriptsize   
\setlength{\tabcolsep}{5pt}
\renewcommand{\arraystretch}{0.95}
\caption{Self- and cross-identity reenactment results.}
\vspace{-3mm}
\begin{tabular}{lcccccc|ccc}
\toprule
& \multicolumn{6}{c|}{\textbf{Self-identity reenactment}}
& \multicolumn{3}{c}{\textbf{Cross-identity reenactment}} \\
\cmidrule(lr){2-7}\cmidrule(lr){8-10}
Method
& PSNR$\uparrow$ & SSIM$\uparrow$ & LPIPS$\downarrow$
& CSIM$\uparrow$ & AED$\downarrow$ & APD$\downarrow$
& CSIM$\uparrow$ & AED$\downarrow$ & APD$\downarrow$ \\
\midrule
LAM
& 18.083 & 0.6673 & 0.2750 & 0.7557 & 0.0337 & 0.2803
& 0.7026 & 0.0466 & 0.5069 \\
Zhang
& 19.310 & 0.7074 & 0.2686 & 0.7629 & 0.0312 & 0.3148
& 0.7077 & 0.0391 & 0.4908 \\
LAM$^\dagger$
& \textbf{20.437} & \textbf{0.7373} & \textbf{0.2315} & \textbf{0.8541} & \textbf{0.0258} & \textbf{0.2272}
& \textbf{0.8111} & \textbf{0.0322} & \underline{0.4245} \\
Zhang$^\dagger$
& \underline{20.003} & \underline{0.7193} & \underline{0.2636} & \underline{0.7679} & \underline{0.0262} & \underline{0.2733}
& \underline{0.7127} & \underline{0.0341} & \textbf{0.4220} \\
\bottomrule
\end{tabular}
\vspace{-8mm}
\label{tab:self_cross_horizontal}
\end{table}
\subsection{Experimental Results}

\subsubsection{Reconstruction and Reenactment Fidelity.}

We evaluate whether explicit emotion control degrades the reconstruction and reenactment performance of the underlying feed-forward backbones. 
We compare LAM~\cite{lam} and Zhang et al.~\cite{zhang2025bringingportrait3dpresence} with their emotion-aware counterparts, LAM$^\dagger$ (LAM$+$Ours) and Zhang$^\dagger$ (Zhang$+$Ours), under both self- and cross-reenactment settings.

Quantitative results on the held-out synthetic test set are reported in Tab.~\ref{tab:self_cross_horizontal}. 
Across image-level (PSNR, SSIM, LPIPS), identity-level (CSIM), and geometry-level (AED, APD, AKD) metrics, our method closely matches the original backbones. 
This indicates that emotion-aware modulation preserves reconstruction fidelity and driving accuracy.

Fig.~\ref{fig:recon_fidelity} shows qualitative comparisons on synthetic data with pixel-aligned ground truth. 
LAM$^\dagger$ and Zhang$^\dagger$ maintain identity consistency, stable geometry, and accurate reenactment comparable to their baselines. 
Additional real-image results are provided in the supplementary material.

\begin{table}[t]
\centering
\scriptsize
\setlength{\tabcolsep}{3.5pt}
\renewcommand{\arraystretch}{1.05}
\caption{Emotion transfer results. Bold/underline denote best/second best.}
\vspace{-3mm}
\begin{tabular}{lcccccc}
\toprule
Metric & P4D-v2 & GAG & LAM & Zhang & LAM$^\dagger$ & Zhang$^\dagger$ \\
\midrule
PSNR$\uparrow$    & 17.684 & 16.965 & 17.434 & 17.762 & \textbf{18.614} & \underline{18.255} \\
SSIM$\uparrow$    & 0.6334 & 0.6399 & 0.6347 & 0.6594 & \textbf{0.6864} & \underline{0.6777} \\
LPIPS$\downarrow$ & 0.3005 & 0.3018 & 0.3052 & 0.3019 & \textbf{0.2678} & \underline{0.2955} \\
CSIM$\uparrow$    & 0.6652 & 0.6616 & 0.6237 & 0.6008 & \textbf{0.7586} & \underline{0.6702} \\
AED$\downarrow$   & 0.0832 & 0.0866 & 0.0735 & 0.0687 & \textbf{0.0314} & \underline{0.0368} \\
APD$\downarrow$   & 0.8373 & 0.7652 & 0.6515 & 0.6572 & \textbf{0.4292} & \underline{0.4859} \\
\bottomrule
\end{tabular}
\vspace{-8mm}
\label{tab:emotion_transfer}
\end{table}

\subsubsection{Emotion Transfer and Disentanglement.}
We next evaluate the effectiveness of explicit emotion control and the degree of disentanglement between emotion, geometry, and appearance.
We compare our method against representative state-of-the-art approaches, including Portrait-4D v2\cite{portrait4dv2}, GAGAvatar\cite{gagavatar}, LAM\cite{lam}, and Zhang et al.\cite{zhang2025bringingportrait3dpresence}.

Quantitative results on the synthetic emotion-transfer benchmark are summarized in Tab.~\ref{tab:emotion_transfer}.
Our method consistently outperforms baselines across all emotion-related metrics. Notably, we achieve a significantly lower Average Expression Distance (AED), which demonstrates that our explicit emotion conditioning enables more accurate expression synthesis aligned with the target emotional state, whereas baselines often remain trapped by the driving signal's original expression.

Fig.~\ref{fig:emotion_transfer} shows qualitative comparisons under emotion transfer. 
While our method produces results consistent with target emotional states, baselines exhibit distinct failure modes in geometry and appearance. 
From a geometric perspective, existing approaches focus on faithful expression reproduction but neglect the underlying emotional semantics, leading to misalignments in mouth shapes or eye shapes when the driving signal conflicts with the target emotion. 
Regarding appearance, feed-forward architectures~\cite{lam, zhang2025bringingportrait3dpresence} suffer from a \textit{baked-in} effect where reference emotional cues (e.g., subtle wrinkles related to specific emotion) are permanently fixed in the texture. In contrast, while optimization-based methods~\cite{portrait4dv2, gagavatar} better reflect the driving signal's expression through per-frame processing, they still fail to pivot the appearance towards the specified target emotion.





\subsubsection{User Study.}

To evaluate perceptual emotion quality on real data, we conduct a user study comparing Portrait-4D v2~\cite{portrait4dv2}, GAGAvatar~\cite{gagavatar}, LAM~\cite{lam}, Zhang et al.~\cite{zhang2025bringingportrait3dpresence}, and our method.

Participants rated results along three dimensions:
image quality, emotion recognizability, and emotional vividness.
The aggregated statistics are reported in Tab.~\ref{tab:user_study}.

\begin{table}[t]
\centering
\scriptsize
\setlength{\tabcolsep}{3.5pt}
\caption{User study results (higher is better).}
\vspace{-3mm}
\label{tab:user_study}
\begin{tabular}{lcccccc}
\toprule
 & P4D-v2 & GAG & LAM & Zhang & LAM$^\dagger$ & Zhang$^\dagger$ \\
\midrule
Image quality & 3.377 & 3.236 & 3.015 & 3.132 & \underline{3.472} & \textbf{3.542} \\
Emotion recognizability & 2.827 & 2.810 & 2.793 & 2.809 & \underline{3.711} & \textbf{3.718} \\
Emotional vividness & 2.965 & 2.954 & 2.926 & 2.968 & \textbf{3.637} & \underline{3.564} \\
Overall & 3.056 & 3.000 & 2.912 & 2.969 & \underline{3.606} & \textbf{3.608} \\
\bottomrule
\end{tabular}
\vspace{-7mm}
\end{table}


While all methods achieve comparable image quality scores, our method consistently ranks highest in both emotion recognizability and emotional vividness. This confirms that explicitly modeling emotion as a controllable and disentangled signal leads to perceptually stronger and more interpretable emotional expressions, even for unseen identities.

\subsection{Ablation Study}
\subsubsection{Impact of Geometry and Appearance Branches.}
We evaluate the contribution of each branch in Fig. \ref{fig:ablation} and Tab. \ref{tab:ablation}. Without the appearance branch, the model degrades to its pre-trained state; it fails to erase the source emotion's texture from the reference image or synthesize target details, such as the dynamic changes in nasolabial folds, forehead wrinkles, and eyebrow direction, as shown in w/o app. of Fig. \ref{fig:ablation}. Conversely, removing the geometry branch retains the reference mesh's emotional geometry, leading to significant deviations from the target geometry in eye opening magnitude and mouth corner movement, as shown in w/o geom. of Fig. \ref{fig:ablation}. Crucially, such inconsistencies between geometry and appearance result in unnatural emotional reproduction, validating the necessity of our dual-branch design.

\subsubsection{Geometry Reusability.}
\begin{figure}[t]
  \centering
  \begin{overpic}[width=0.8\linewidth]{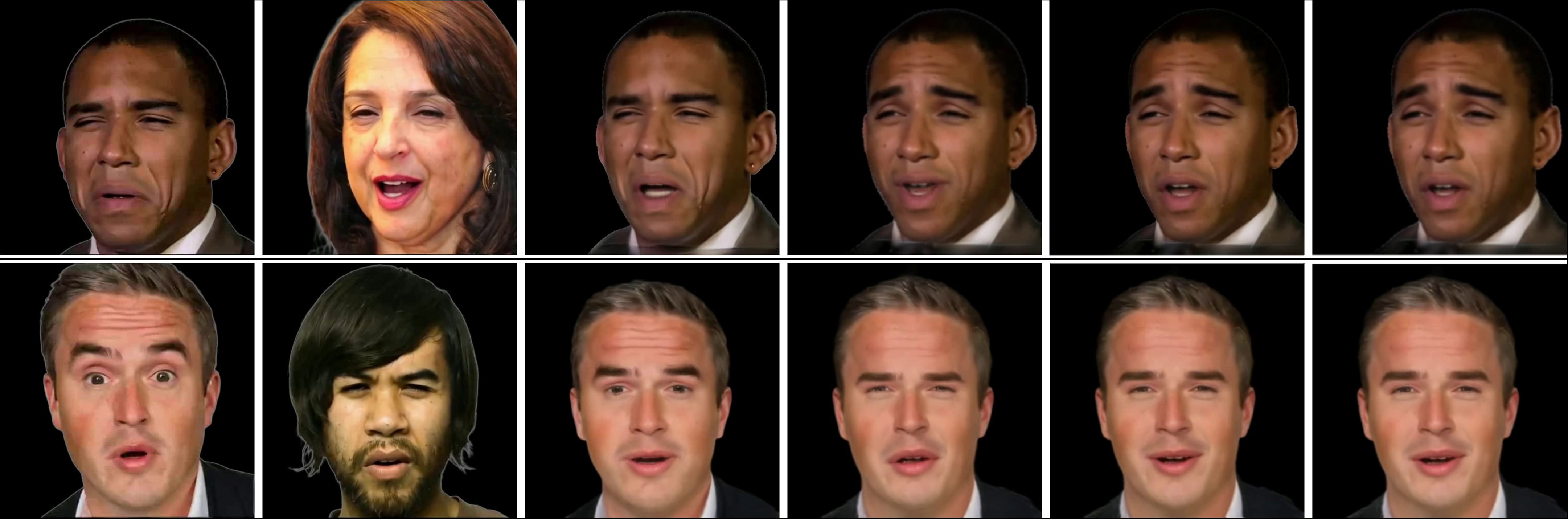}
    \put(5.5,-2.5){\footnotesize Ref.}
    \put(-2,14){%
    \setlength{\fboxsep}{2pt}
    \setlength{\fboxrule}{0.4pt}
    \fcolorbox{black}{white!50}{\tiny Surprised}
  }
    \put(-2,30.7){%
      \setlength{\fboxsep}{2pt}
    \setlength{\fboxrule}{0.4pt}
    \fcolorbox{black}{white!50}{\tiny Disgust}
    }
    
    \put(22,-2.5){\footnotesize Drv.}
    \put(14.7,14){%
    \setlength{\fboxsep}{2pt}
    \setlength{\fboxrule}{0.4pt}
    \fcolorbox{black}{white!50}{\tiny Angry}
  }
  \put(14.7,30.7){%
    \setlength{\fboxsep}{2pt}
    \setlength{\fboxrule}{0.4pt}
    \fcolorbox{black}{white!50}{\tiny Happy}
  }
    
    \put(35,-2.5){\footnotesize w/o app.}
    \put(50.5,-2.5){\footnotesize w/o geom.}
    \put(72,-2.5){\footnotesize Full}
    \put(84,-2.5){\footnotesize reuse geom.}
    \put(65,14){%
    \setlength{\fboxsep}{2pt}
    \setlength{\fboxrule}{0.4pt}
    \fcolorbox{black}{white!50}{\tiny Happy}
  }
  \put(65,30.7){%
    \setlength{\fboxsep}{2pt}
    \setlength{\fboxrule}{0.4pt}
    \fcolorbox{black}{white!50}{\tiny Sad}
  }
    
    \put(-3.5,20){\rotatebox{90}{\footnotesize LAM$^\dagger$}}
    \put(-3.5,3){\rotatebox{90}{\footnotesize Zhang$^\dagger$}}
  \end{overpic}
  \captionof{figure}{Ablation of dual-path emotion modulation. 
Without appearance modulation, texture-level emotion leaks from the reference; without geometry modulation, deformation follows the driving motion. 
The full model suppresses both and matches the target emotion.
Additionally, geometry reuse across backbones achieves comparable results, highlighting its robust transferability.}
  \label{fig:ablation}
  \vspace{-3mm}
\end{figure}
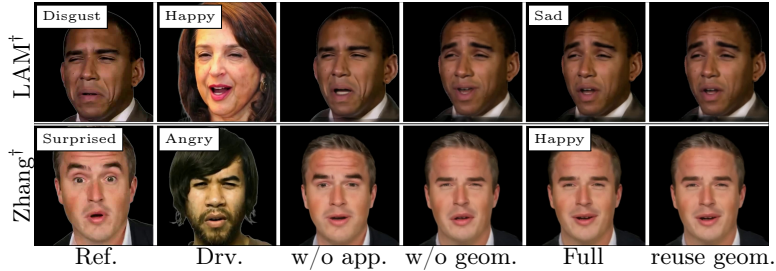

To evaluate the generality of the geometry branch, we investigate the reusability of the inferred geometry sequences. Since both LAM$^\dagger$ and Zhang$^\dagger$ operate within the same canonical FLAME parameter space, the geometry branch can be seamlessly reused across architectures. Specifically, we employ the geometry branch from Zhang$^\dagger$ to train the appearance branch of LAM$^\dagger$. As shown in Fig.~\ref{fig:ablation}, performance remains consistently strong after this geometry reuse, suggesting good cross-backbone transferability. 


\begin{table}[t]
\centering
\scriptsize
\setlength{\tabcolsep}{3.2pt}
\renewcommand{\arraystretch}{0.96}
\caption{Ablation study.}
\label{tab:ablation}
\vspace{-4mm}
\resizebox{\linewidth}{!}{
\begin{tabular}{lccccccc}
\toprule
& \multicolumn{3}{c}{\textbf{LAM + Ours}} & \multicolumn{3}{c}{\textbf{Zhang et al. + Ours}} & \textbf{Reuse geom.} \\
\cmidrule(lr){2-4} \cmidrule(lr){5-7}
Metric & w/o app. & w/o geom. & Full & w/o app. & w/o geom. & Full & \\
\midrule
PSNR$\uparrow$   & 18.195 & \underline{18.422} & \textbf{18.614} & 17.958 & \underline{18.118} & \textbf{18.255} & 18.515 \\
SSIM$\uparrow$   & 0.6659 & \underline{0.6669} & \textbf{0.6864} & \underline{0.6635} & 0.6603 & \textbf{0.6777} & 0.6878 \\
LPIPS$\downarrow$& 0.2778 & \underline{0.2762} & \textbf{0.2678} & \underline{0.2957} & 0.3019 & \textbf{0.2955} & 0.2819 \\
CSIM$\uparrow$   & \underline{0.7234} & 0.7181 & \textbf{0.7586} & \underline{0.6383} & 0.6227 & \textbf{0.6702} & 0.6724 \\
\bottomrule
\end{tabular}}
\vspace{-6mm}
\end{table}

\section{Discussion}
\subsection{Emotion Control Analysis}
\subsubsection{Emotion Transfer Across Identities.}
This experiment evaluates emotion transfer across identities by fixing both the target emotion and the driving geometry, while varying the input identity provided by the reference image. This setting isolates identity-dependent appearance responses under a shared emotional state and identical facial motion.

\begin{figure}[t]
    \centering
  \begin{overpic}[width=0.9\linewidth]{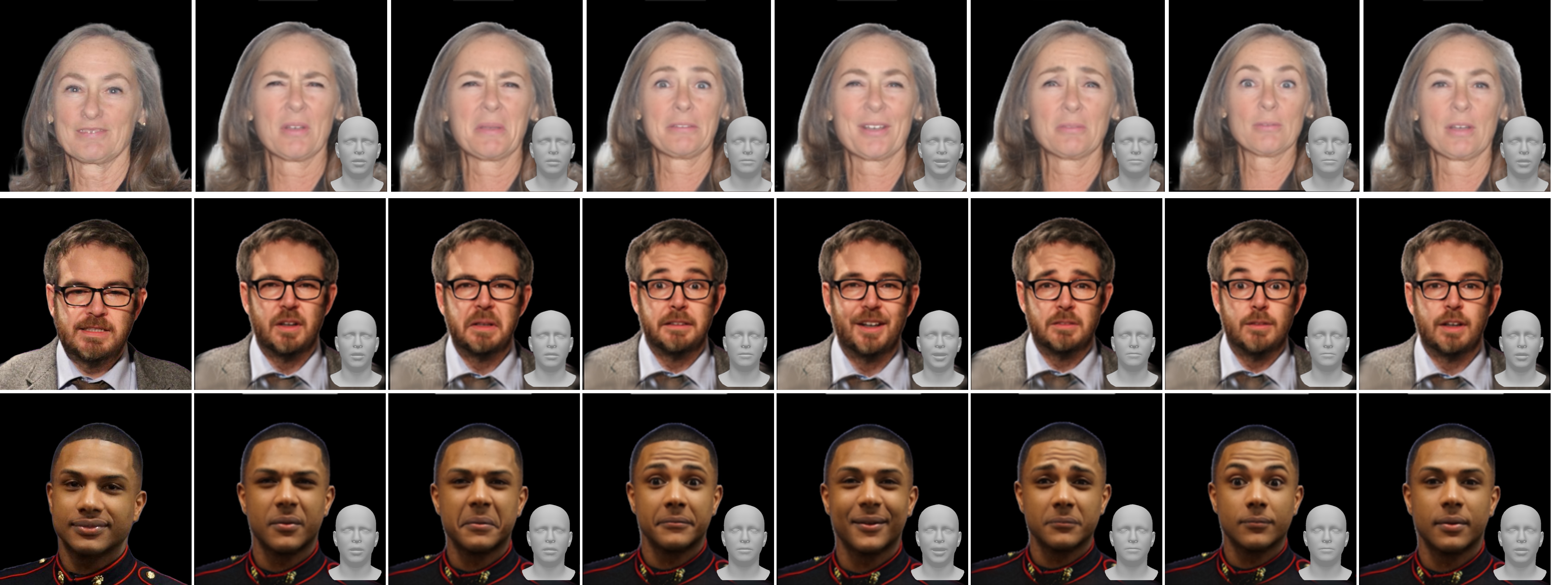}
    \put(4,-2.5){\footnotesize Ref.}
    \put(14.5,-2.5){\footnotesize Angry}
    \put(26,-2.5){\footnotesize Disgust}
    \put(40.5,-2.5){\footnotesize Fear}
    \put(51.5,-2.5){\footnotesize Happy}
    \put(65.5,-2.5){\footnotesize Sad}
    \put(74.5,-2.5){\footnotesize Surprised}
    \put(88,-2.5){\footnotesize Neutral}
  \end{overpic}
\captionof{figure}{Emotion control across identities under fixed geometry. 
Rows: identities; columns: seven target emotions. 
All results share the same driving FLAME sequence. 
Identities respond differently to the same emotion, while each identity exhibits consistent emotion-specific variation.}
  \label{fig:discussion_1_2}
  \vspace{-8mm}
\end{figure}
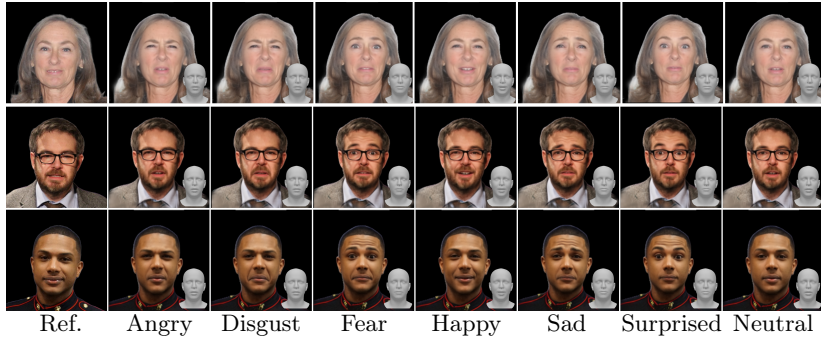

As shown in Fig.~\ref{fig:discussion_1_2}, the four identities differ in gender, age, and skin condition. Despite sharing the same driving geometry and emotion control signal, the synthesized avatars exhibit distinct and plausible appearance variations. In particular, our model generates dynamic texture details, such as wrinkles and skin tension, that are visually consistent with each subject’s identity rather than following a single, shared emotion-specific pattern.

These results indicate that the proposed model does not represent emotion as a fixed appearance template applied uniformly across identities. Instead, it learns an identity-conditioned emotion distribution, enabling the same emotional state and motion to manifest differently across individuals. This behavior highlights the effectiveness of our appearance modulation design in capturing identity-aware emotional cues under explicit emotion control.

\subsubsection{Emotion Consistency Under Fixed Geometry.}
This experiment evaluates emotion control under a fixed driving geometry by varying the target emotion while keeping both the motion sequence and identity unchanged. Specifically, we start from a tracked FLAME sequence corresponding to the same utterance and content, and apply different emotion controls through the geometry modulation branch.

As illustrated in Fig.~\ref{fig:discussion_1_2}, the emotion-specific geometry predicted by our model exhibits clear and systematic deviations that are consistent with the intended emotional states. For example, \textit{happy} results in lifted mouth corners, \textit{disgust} produces downturned mouth corners with tightened brows, and \textit{surprised} leads to widened eyes. Importantly, these emotion-dependent deformations are consistently applied on top of the same underlying articulation, preserving lip motion and speech-related dynamics.

The results further show that identity characteristics remain stable across emotion changes, with no observable identity drift. Together, these observations demonstrate that the geometry modulation branch enables explicit and controllable emotion variation while preserving articulation and speech-related motion.
\subsubsection{Motion Robustness Under Fixed Emotion.}
\begin{figure}[t]
  \centering
  \includegraphics[width=0.7\linewidth]{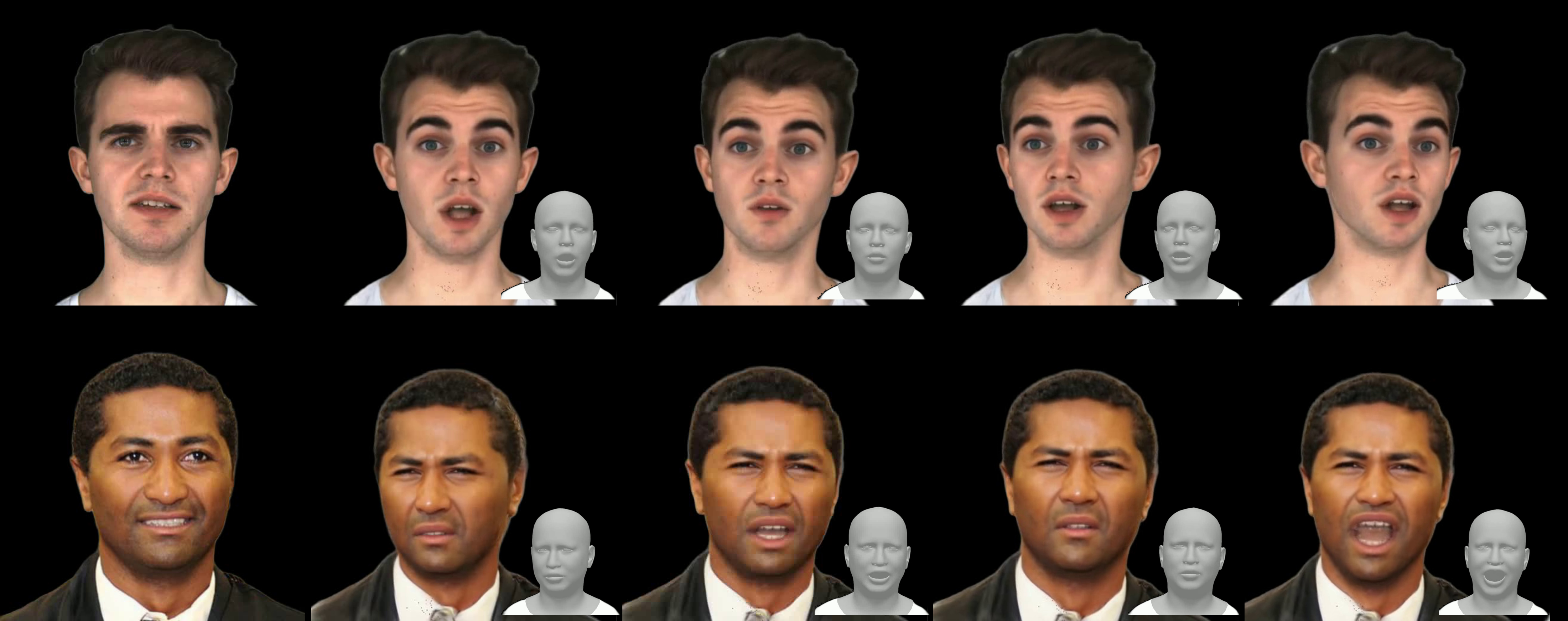}
  \vspace{-2mm}
\captionof{figure}{Motion robustness under fixed emotion. 
Rows fix identity and emotion; columns vary driving FLAME motion. 
Emotion remains consistent across articulations while preserving identity.}
  \label{fig:discussion_3}
  \vspace{-4mm}
\end{figure}

Our dataset construction provides comprehensive emotion-conditioned variations while preserving identity, which allows our model to
transfer emotions to a fixed identity without altering its identity cues.
As shown in Fig.~\ref{fig:discussion_3}, all identities maintain high-quality identity consistency after reenacting different target emotions. Additional visual results are provided in the supplementary material. 
We also observe identity-dependent differences in emotion prediction across subjects, which we discuss in the Discussion section.

\subsection{Emotion Interpolation}

Although our model is trained using a \emph{finite} set of seven discrete emotion labels, it interestingly generalizes to \emph{continuous} emotion mixtures at inference time.
Specifically, we interpolate between emotion embeddings in the learned emotion space and use the resulting embedding to condition both the geometry and appearance modulation branches.
No interpolation is performed directly on FLAME parameters, motion sequences, or rendered attributes.

\begin{figure}[t]
    \centering
  \includegraphics[width=0.8\linewidth]{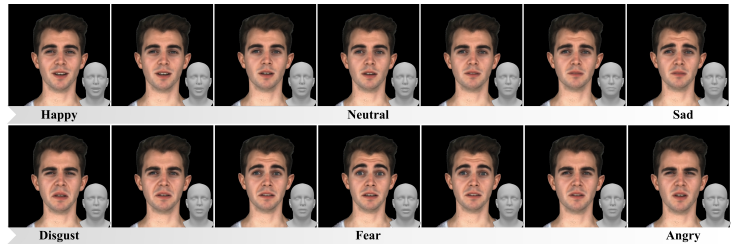}
  \vspace{-3.5mm}
\captionof{figure}{Continuous emotion interpolation. 
For a fixed identity, rows interpolate between \emph{happy–neutral–sad} and \emph{disgust–fear–angry}. 
Despite discrete training labels, embedding interpolation yields smooth geometry and appearance transitions. 
FLAME geometry is shown in the inset.}
  \label{fig:emo_interpola}
  \vspace{-8mm}
\end{figure}

As shown in Fig.~\ref{fig:emo_interpola}, smoothly interpolating the emotion embedding from \textit{happy} to \textit{neutral} and further to \textit{sad} produces a perceptually continuous transition in affect.
Throughout the interpolation, identity-specific facial characteristics remain stable, speech-driven articulation is preserved, and no temporal artifacts or expression collapse are observed.
Both geometric deformation and appearance cues evolve consistently with the interpolated emotional state, indicating that the learned emotion representation forms a smooth and structured control space beyond the discrete training categories.

\section{Conclusion and Limitations}

We presented a plug-and-play framework for explicit emotion control in feed-forward single-image 3D head avatar reconstruction. By modeling emotion as a first-class, identity-agnostic signal, our method decomposes emotional variation into geometry normalization and identity-aware appearance modulation while remaining compatible with existing reconstruction pipelines. Supported by a time-synchronized, multi-identity dataset, the framework enables controllable and identity-preserving emotion transfer without compromising reconstruction fidelity. Experiments across multiple backbones demonstrate improved expressiveness, disentanglement, and controllability, validating the effectiveness of explicit emotion modeling for scalable 3D avatars.

Despite its advantages, our approach has limitations. First, the reliance on emotion-synchronized anchor sequences constrains the diversity of speech content and emotional dynamics; extending synchronization strategies to broader settings is future work. Second, performance degrades under large yaw or near-profile views due to limited extreme-pose supervision in the training data.
\putbib[main]
\end{bibunit}

\clearpage

\setcounter{section}{0}
\setcounter{figure}{0}
\setcounter{table}{0}
\setcounter{equation}{0}
\setcounter{page}{1}

\renewcommand{\thesection}{S\arabic{section}}
\renewcommand{\thefigure}{S\arabic{figure}}
\renewcommand{\thetable}{S\arabic{table}}
\renewcommand{\theequation}{S\arabic{equation}}

\makeatletter
\makeatother

\begin{center}
    {\Large \bfseries Giving Faces Their Feelings Back \par}
    \vskip .2cm
    {\large Explicit Emotion Control for Feedforward Single-Image 3D Head Avatars \par}
    \vskip .2cm
    {\large --- Supplementary Materials --- \par}
    \vskip .8cm
\end{center}

\begin{bibunit}

\section{Implementation Details}

\subsection{Training Objectives}
\label{sec:training_objectives}

We train the geometry modulation branch and the appearance modulation branch jointly with separate objectives, using synchronized emotion supervision.

\paragraph{Geometry modulation loss.}
The geometry branch takes a speech-driven deformation sequence and a target emotion label, and predicts an emotion-aligned deformation sequence using the original parametric face model.
We supervise this branch using losses defined in the parameter and surface domains, enforcing consistency with the ground-truth deformation sequence of the same content and target emotion:
\begin{equation}
\mathcal{L}_{\mathrm{geo}}
=
\lambda_{\mathrm{param}}
\|\tilde{\mathbf{p}} - \mathbf{p}^{\ast}\|_2^2
+
\lambda_{\mathrm{surf}}
\|\tilde{\mathbf{V}} - \mathbf{V}^{\ast}\|_2^2,
\end{equation}
where $\tilde{\mathbf{p}}$ and $\tilde{\mathbf{V}}$ denote the predicted deformation parameters and the corresponding mesh vertices, and $(\mathbf{p}^{\ast}, \mathbf{V}^{\ast})$ are the ground-truth values under the same articulation and target emotion.

\paragraph{Appearance modulation loss.}
The appearance branch injects an explicit emotion signal into the reconstruction pipeline to synthesize emotion-aware appearance.
Given a reference image and a target emotion label, we supervise the final rendered output against the corresponding target-emotion frame using a reconstruction loss:
\begin{equation}
\mathcal{L}_{\mathrm{app}}
=
\|\tilde{I} - I^{\ast}\|_2^2,
\end{equation}
where $\tilde{I}$ is the synthesized image and ${I}^{\ast}$ is the ground-truth frame with the same identity, content, and target emotion.

\subsection{Dataset Construction and Statistics}

This section provides detailed dataset statistics and construction procedures omitted from the main paper for clarity.

\subsubsection{Synthesis--Tracking Pipeline.}
All synthetic data are generated using our synthesis pipeline. Each video is rendered with known camera parameters and subsequently processed by a unified tracking system to extract consistent FLAME\cite{flame} parameters (including expression, jaw pose, and head pose). This ensures that both real and synthetic data share a common parametric representation.

All datasets used for quantitative and qualitative evaluation are strictly held out from training. In particular, anchor emotion sequences used for emotion synchronization are never used during evaluation.

\subsubsection{Appearance Configuration.}
The appearance configuration is designed to emphasize identity diversity and appearance coverage, while placing a weaker constraint on temporal continuity.

We sample videos at 5\,fps. A total of 8{,}750 reference identities are used. For each identity, we select a reference clip whose viewpoint is closest to the reference image, and render reference videos under seven expression conditions. This results in 61{,}250 appearance-focused videos in total.

\subsubsection{Geometry Configuration.}
The geometry configuration prioritizes temporal coherence and stable motion patterns for learning geometry and expression modulation.

Videos are sampled at 30\,fps. We generate 8{,}750 videos from 1{,}250 reference identities under the same synthesis settings as the appearance configuration. Compared to the appearance dataset, this configuration trades identity diversity for longer and temporally consistent motion sequences.

\subsubsection{Additional Real and Synthetic Data.}
To support training and evaluation of Zhang et al.~\cite{zhang2025bringingportrait3dpresence}, we additionally incorporate real facial video datasets (e.g., VFHQ\cite{xie2022vfhq}) and synthetic data produced using their original pipeline. All additional data are processed through the same tracking system to ensure consistent FLAME parameterization across sources.

\subsection{Architecture and Training Details}

\subsubsection{Emotion Representation.}
Emotion labels are encoded as discrete one-hot vectors. Each non-neutral emotion is represented by a 6D one-hot vector, while the neutral emotion is mapped to the all-zero vector. Positional embeddings are applied, and the resulting representation is projected to a 1024-dimensional embedding whose length matches the Transformer token sequence.

The emotion embedding is concatenated with image tokens along the token dimension and provided as conditioning input to the Transformer.

\subsubsection{Geometry Modulation Network.}
The geometry modulation module adopts the same 4-layer Transformer backbone as used in LAM~\cite{lam}. Its input is a 103-dimensional vector formed by concatenating FLAME expression and jaw pose parameters. The module outputs emotion-conditioned geometry features that modulate the driving FLAME sequence.

The loss weight for geometry modulation is set to $\lambda = 1$ for all experiments.

\subsubsection{Training Settings.}
For both LAM\cite{lam} and Zhang et al.~\cite{zhang2025bringingportrait3dpresence}, we follow the original training protocols to ensure fair comparison. This includes reference-image preprocessing, FLAME upsampling ratio, rendering resolution, optimizer choice, batch size, and learning-rate schedules.

When fine-tuning from released checkpoints (e.g., LAM~\cite{lam}), all weights are initialized from the official models. For our implementation of Zhang et al.~\cite{zhang2025bringingportrait3dpresence}, we replace the template mesh of FlashAvatar~\cite{flashavatar} with FLAME while keeping the remaining architecture unchanged. Detailed training configurations, including hardware resources, iterations, and total durations, are summarized in Table~\ref{tab:training_efficiency}. Additional implementation details and hyper-parameters are kept identical across baseline and emotion-aware variants unless explicitly stated.

\begin{table}[H]
    \centering
    \caption{Comparison of training configurations and computational efficiency.}
    \label{tab:training_efficiency}
    \begin{threeparttable}
        \begin{tabularx}{\linewidth}{@{} *{4}{>{\centering\arraybackslash}X} @{}}
            \toprule
            Method & Hardware & Iterations & Training Time \\
            \midrule
            LAM~\cite{lam} & 8 $\times$ A100 (80G)\tnote{*} & 45,500\tnote{**} & $\sim$2 weeks\tnote{*} \\
            Zhang et al.~\cite{zhang2025bringingportrait3dpresence} & 8 $\times$ A100 (80G) & 250,000 & 42.6 h \\
            LAM$^\dagger$ & 4 $\times$ A100 (80G) & 30,000 & 58.4 h \\
            Zhang$^\dagger$ & 4 $\times$ A100 (80G) & 50,000 & 14.8 h \\
            \bottomrule
        \end{tabularx}
        \begin{tablenotes}
            \footnotesize
            \item[*] Training statistics for the original LAM are sourced from official GitHub repository issues.
            \item[**] Inferred from official LAM-20K checkpoint filenames.
        \end{tablenotes}
    \end{threeparttable}
\end{table}



\section{Additional Evaluation Details}

\subsection{Held-Out Evaluation Protocol}
All identities, sequences, and motions used for quantitative and qualitative evaluation are unseen during training. In particular, no anchor emotion sequence used for emotion synchronization appears in the evaluation set. This prevents information leakage and ensures fair assessment of generalization.

\subsection{User Study Details}
We conduct a user study to evaluate the quality of portraits by comparing results produced by different 3D head modeling methods against real data. Specifically, we sample 30 videos from the test set, covering diverse identities, viewpoints, and emotions. We invite 141 participants to rate these videos on a scale from 1 (worst) to 5 (best) with respect to three aspects: image quality, emotion recognizability, and emotion vividness. The evaluation criteria for these metrics are as follows:
\begin{itemize}
    \item \textbf{Image quality}: Evaluate the visual fidelity and overall realism of the videos. Participants are instructed to look for visual artifacts, blurring, unnatural distortions, and lighting/texture inconsistencies. A score of 1 indicates severe visual artifacts and a highly artificial appearance, whereas a score of 5 represents photorealistic quality that is visually indistinguishable from real video footage.
    \item \textbf{Emotion recognizability}: Assess how accurately and easily the intended emotional state can be identified from the facial expressions. Participants should consider whether the target emotion is conveyed without ambiguity. A score of 1 means the emotion is completely unrecognizable or confusing, while a score of 5 indicates that the target emotion is instantly and unambiguously clear.
    \item \textbf{Emotion vividness}: Judge the naturalness, intensity, and dynamic richness of the portrayed emotion. This criterion focuses on the presence of realistic micro-expressions, smooth temporal transitions, and human-like expressiveness rather than just static shapes. A score of 1 reflects stiff, robotic, or rigidly artificial expressions, whereas a score of 5 denotes highly nuanced, authentic, and dynamically convincing emotional performances.
\end{itemize}

\subsection{Additional Results}
The supplemental material includes extended qualitative comparisons for self-reenactment, cross-reenactment, and emotion transfer on both synthetic and real images, following the same visualization protocol as the main paper.
\subsubsection{Reconstruction and reenactment fiedality.}
\begin{figure}[htbp]
    \centering
    \begin{minipage}{\linewidth}
    \begin{overpic}[width=\linewidth]{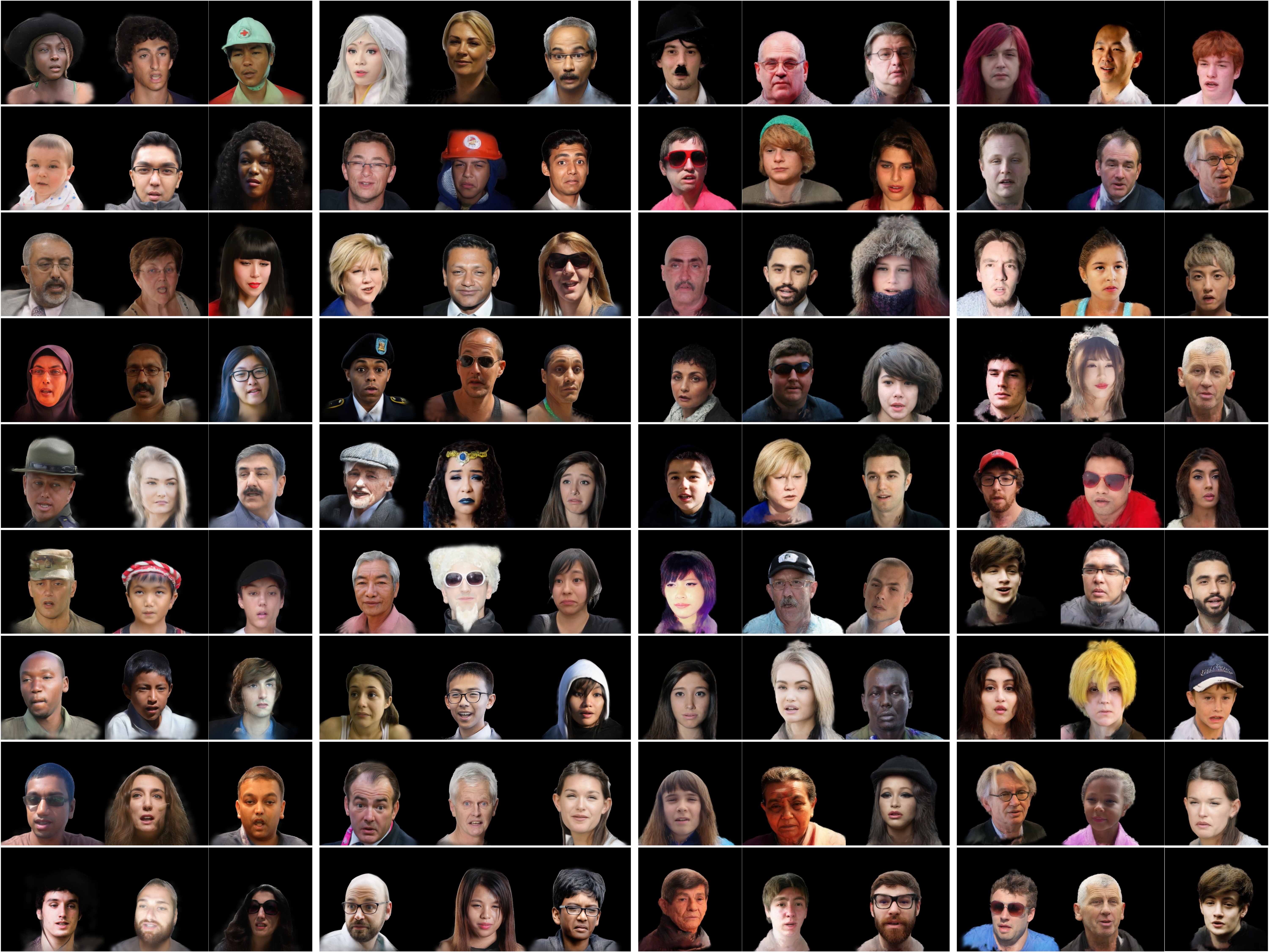}

    \put(7,-3){ LAM}
    \put(35,-3){LAM$^\dagger$}
    \put(59,-3){Zhang}
    \put(85.5,-3){Zhang$^\dagger$}
    
    \end{overpic}
    \end{minipage}
    \caption{Reconstruction fidelity galleries on real-world data.}
    \label{fig:recon_fidelity_galleries}
\end{figure}

Fig. \ref{fig:recon_fidelity_galleries} provides a gallery-style visualization of the reconstruction and reenactment results. Notably, for our methods (LAM$^\dagger$ and Zhang$^\dagger$), we additionally present the visual results under different ablation settings. The qualitative comparisons demonstrate that our approach does not compromise rendering quality compared to the baselines.

\subsubsection{Emotion transfer}
\begin{figure}[htbp]
    \centering
    \begin{minipage}{\linewidth}
    \begin{overpic}[width=\linewidth]{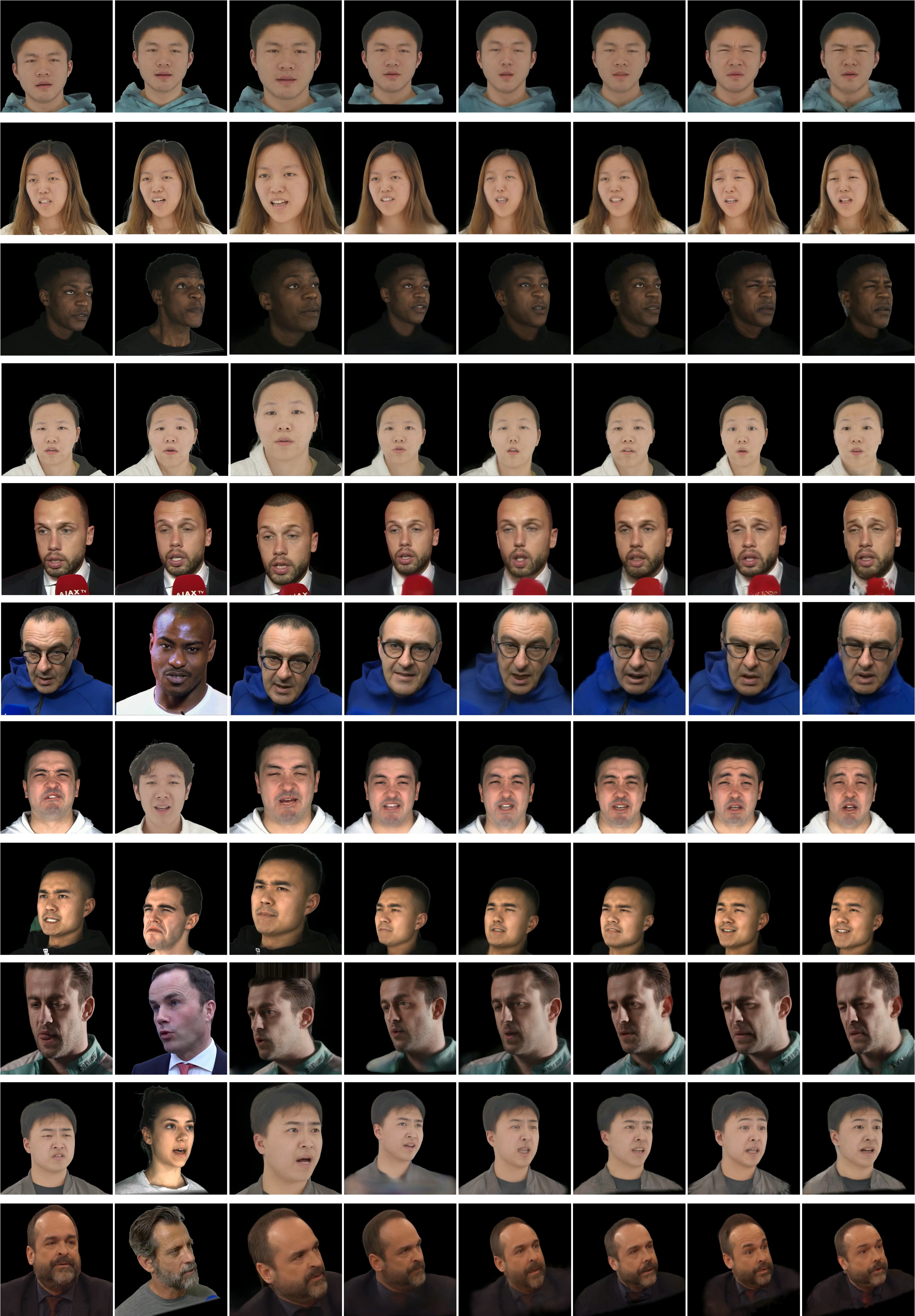}
    \put(3,-1.5){\footnotesize Ref.}
    \put(11.5,-1.5){\footnotesize Drv.}
    \put(18.5,-1.5){\footnotesize P4D-v2}
    \put(28.5,-1.5){\footnotesize GAG}
    \put(37,-1.5){\footnotesize LAM}
    \put(45.5,-1.5){\footnotesize Zhang}
    \put(54,-1.5){\footnotesize LAM$^\dagger$}
    \put(62.5,-1.5){\footnotesize Zhang$^\dagger$}

    \put(-1.5,93){\rotatebox{90}{\footnotesize Angry}}
    \put(-1.5,84.5){\rotatebox{90}{\footnotesize Sad}}
    \put(-1.5,74){\rotatebox{90}{\footnotesize Disgust}}
    \put(-1.5,64){\rotatebox{90}{\footnotesize Surprised}}
    \put(-1.5,57){\rotatebox{90}{\footnotesize Sad}}
    \put(-1.5,47){\rotatebox{90}{\footnotesize Angry}}
    \put(-1.5,39){\rotatebox{90}{\footnotesize Sad}}
    \put(-1.5,29){\rotatebox{90}{\footnotesize Happy}}
    \put(-1.5,19){\rotatebox{90}{\footnotesize Disgust}}
    \put(-1.5,11){\rotatebox{90}{\footnotesize Fear}}
    \put(-1.5,0.5){\rotatebox{90}{\footnotesize Surprised}}
    
    \end{overpic}
    
    \end{minipage}
    \caption{Emotion transfer on real-world data.}
    \label{fig:emo_transfer_real}
\end{figure}

Fig. \ref{fig:emo_transfer_real} further demonstrates the robustness of our model for emotion transfer on real-world data. Specifically, we evaluate the performance under two distinct settings: an intra-identity setting (the upper 5 rows), where the reference image and driving video share the same identity, and a cross-identity setting (the lower 6 rows), where the identities differ. It is worth noting that the emotion transfer task is inherently identity-agnostic. The core challenge lies in the effective disentanglement and removal of the source emotion, coupled with the accurate reconstruction of the target emotion dictated by the emotion label. Experimental results demonstrate that our proposed method successfully eliminates the source emotion and faithfully synthesizes the target emotion under both settings.


\subsubsection{Ablation study.}
\begin{figure}[t]
  \centering
  \begin{overpic}[width=\linewidth]{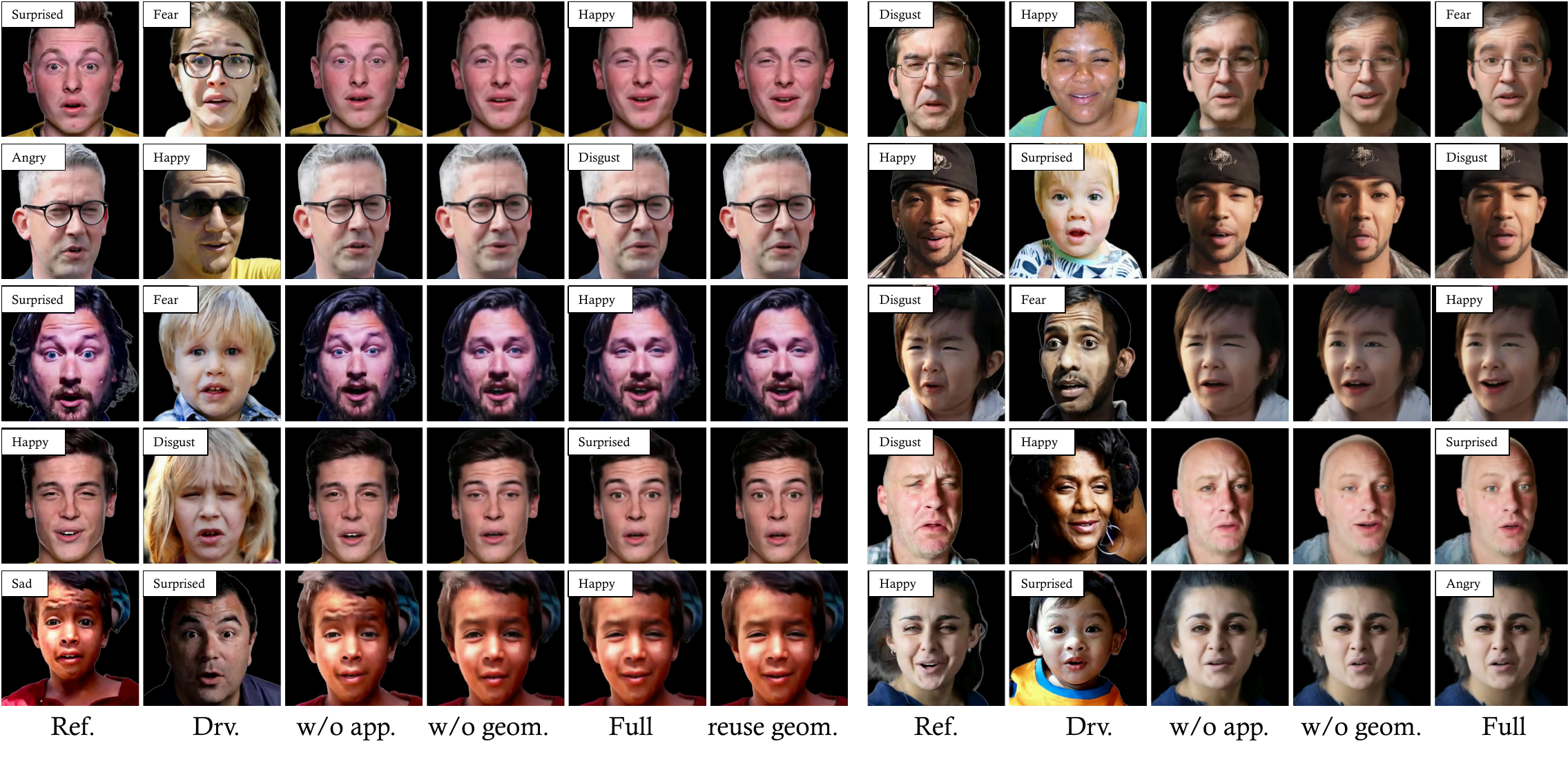}
    
    
    
  \end{overpic}
  \captionof{figure}{Additional results of ablation study.}
  \label{fig:supp_ablation}
\end{figure}

Fig. \ref{fig:supp_ablation} shows additional results of ablation study.  The qualitative comparisons demonstrate that ablating any individual module leads to a degradation in rendering fidelity. Specifically, omitting the geometry branch introduces geometric inconsistencies (e.g., inaccuracies in the degree of eye opening, and the elevation or depression of the mouth corners), while removing the appearance branch causes textural inconsistencies (primarily manifested as the unnatural appearance or disappearance of wrinkles).

\subsection{Continuous Emotion Interpolation}

\begin{figure}[htbp!]
    \centering
    \begin{minipage}{\linewidth}
    \includegraphics[width=\linewidth]{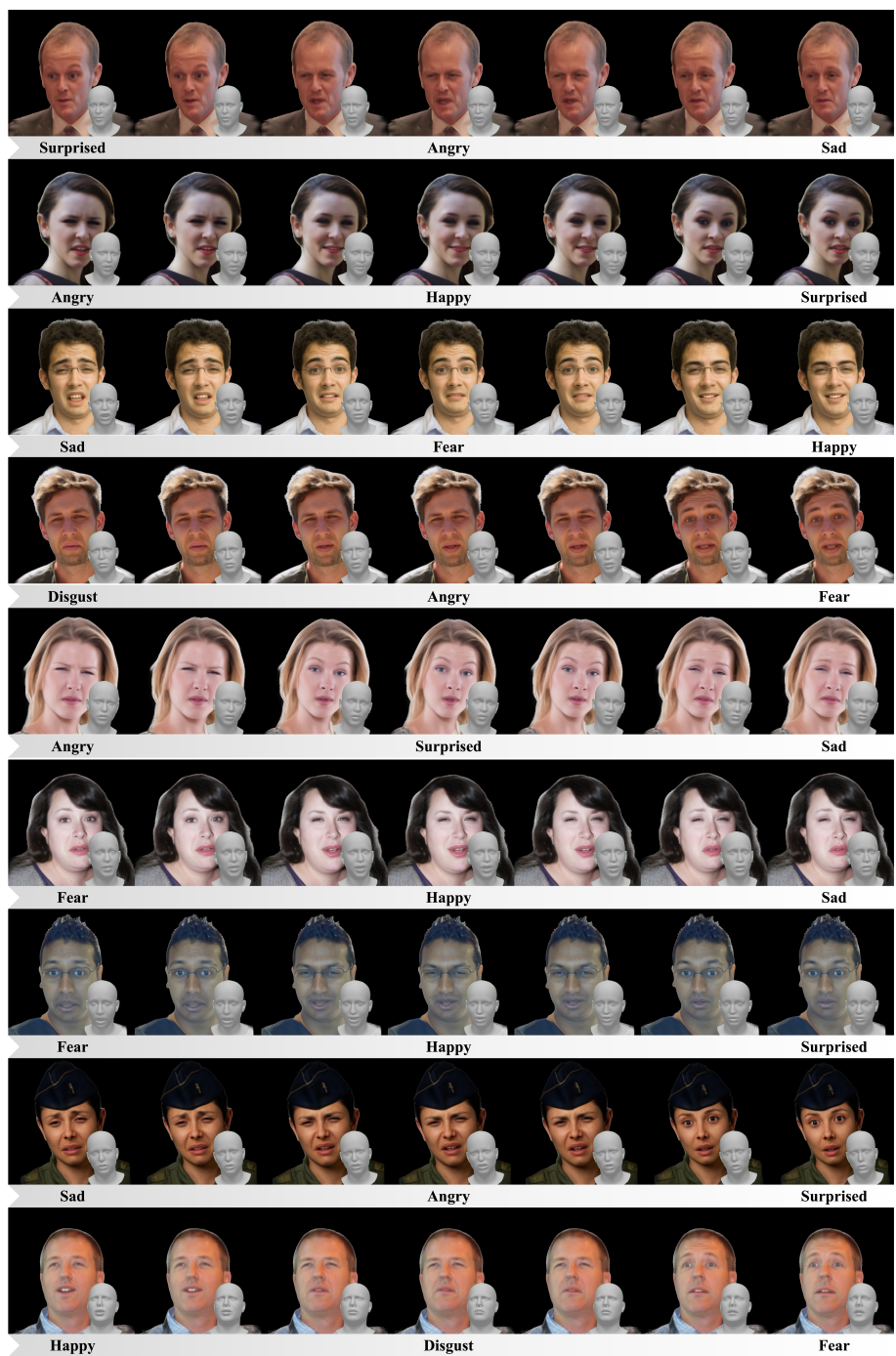}
    \end{minipage}
    \caption{Emotion interpolation on real-world data.}
    \label{fig:emo_inter_real}
\end{figure}

To further demonstrate the generalization capability of our model, we provide additional visualization of emotion interpolation in Fig.~\ref{fig:emo_inter_real}. Although the network is supervised using only seven discrete emotion categories, the results suggest that the model successfully learns a structured and continuous emotion manifold.

Rather than performing a naive linear blend of output parameters (such as FLAME coefficients or motion sequences), which often leads to "expression collapse" or non-physical distortions, our method interpolates within the learned emotion embedding space. This latent representation effectively conditions the geometry deformation and appearance modulation branches in a holistic manner. As illustrated in Fig.~\ref{fig:emo_inter_real}, by traversing the path between distinct emotion poles (e.g., from \textit{angry} to \textit{neutral}, or \textit{surprised} to \textit{happy}), our model generates a spectrum of intermediate affective states that were never explicitly present in the training set.

Throughout these continuous transitions, several key properties are maintained:
\begin{itemize}
\item \textbf{Identity Decoupling:} The fine-grained geometric details and textural features that define the individual's identity remain remarkably stable, regardless of the emotional intensity.
\item \textbf{Motion Compatibility:} The interpolated emotional expressions naturally coexist with speech-driven lip movements, confirming that our emotion conditioning is well-disentangled from the primary articulation stream.
\item \textbf{Geometric-Appearance Synergy:} Both the subtle skin wrinkling (appearance) and the macro-level facial muscle contractions (geometry) evolve in perfect synchrony along the interpolation path.
\end{itemize}
These extended results further validate that our learned emotion space is not merely a set of discrete anchors, but a smooth, navigable domain capable of high-fidelity, fine-grained affective control for 3D head avatars.

\newpage

\putbib[main]
\end{bibunit}

\end{document}